%% file: sn-article.tex
\documentclass[12pt]{article}

\usepackage{graphicx, multirow, amsmath, amssymb, amsfonts, amsthm, mathrsfs}
\usepackage[title]{appendix}
\usepackage{xcolor, booktabs, algorithm, algpseudocode, listings, xurl, float}
\usepackage{tabularx, subcaption, array, caption, rotating}
\usepackage{fontspec}
\newfontfamily\bengalifont[Script=Bengali, AutoFakeBold=1.5, Path=./, Extension=.ttf]{kalpurush}

\usepackage[hidelinks]{hyperref}

\theoremstyle{plain}

\theoremstyle{definition}

\theoremstyle{remark}

\graphicspath{{images/}}

\title{\textbf{Advancing Bangla Machine Translation Through Informal Datasets}}

\author{
Ayon Roy\thanks{ext.ayon.roy@bracu.ac.bd}, 
Risat Rahaman\thanks{risat.rahaman@g.bracu.ac.bd}, 
Sadat Shibly\thanks{sadat.noor.shibly@g.bracu.ac.bd},\\
Udoy Saha\thanks{udoy.saha.joy@g.bracu.ac.bd}, 
Abdulla Al Kafi\thanks{abdulla.al.kafi@g.bracu.ac.bd}, 
Farig Yousuf Sadeque\thanks{farig.sadeque@bracu.ac.bd}\\[1ex]
\small Department of Computer Science and Engineering, BRAC University, Dhaka, Bangladesh
}

\begin{document}

\maketitle

\begin{abstract}
Bangla is the sixth most widely spoken language globally, with approximately 234 million native speakers. However, progress in open-source Bangla machine translation remains limited. Most online resources are in English and often remain untranslated into Bangla, excluding millions from accessing essential information. Existing research in Bangla translation primarily focuses on formal language, neglecting the more commonly used informal language. This is largely due to the lack of pairwise Bangla-English data and advanced translation models. If datasets and models can be enhanced to better handle natural, informal Bangla, millions of people will benefit from improved online information access. In this research, we explore current state-of-the-art models and propose improvements to Bangla translation by developing a dataset from informal sources like social media and conversational texts. This work aims to advance Bangla machine translation by focusing on informal language translation and improving accessibility for Bangla speakers in the digital world.
\end{abstract}


\section{Introduction}
\input{chapters/chapter_1}

\section{Literature Review}
\input{chapters/chapter_2}

\section{Dataset}
\input{chapters/chapter_3}

\section{Methodology}
\input{chapters/chapter_4}

\section{Results and Analysis}
\input{chapters/chapter_5}

\section{Conclusion and Future Work}
\input{chapters/chapter_6}

\section*{Acknowledgements}
The authors wish to express their sincere gratitude for the support received throughout this research. Special thanks to Dr. Farig Yousuf Sadeque (Associate Professor, Department of CSE, BRAC University).

\section*{Declarations}
\begin{itemize}
\item No participant will be hurt with the language and sentences of the dataset.
\item Data is available on Hugging Face: \url{https://huggingface.co/datasets/AyonRoy29/informal_bn-en_machine_translation_dataset}
\end{itemize}

\bibliographystyle{plain} 
\bibliography{submission-refs}

\end{document}

%% file: chapters/chapter_1.tex
The language most commonly used in daily interactions is often colloquial and informal, and is much more ubiquitous than the formal register used in official settings, journals, and technical articles. This is a major handicap for the Bangla language, which is spoken by some 234 million native speakers in regions including Bangladesh and India. Bangla is commonly considered as a low-resource language (Bhattacharjee et al, 2021)\cite{bhattacharjee2021banglabert}, mostly due to the lack of high-quality parallel data required for training robust machine translation (MT) systems.

\subsection{Background}
Traditional MT research and development has primarily focused on translating formal language, which was supported by the availability of resources such as academic and governmental sources. The possibility to use the more common informal language effectively has thus been overlooked due to this strong emphasis (Islam et al, 2010; Baziotis et al, 2020)\cite{islam2010english}\cite{baziotis2020language}. The informal dataset is far less accessible than its formal version, which exacerbates the fundamental issue discussed in this study. \\[5pt]\

Furthermore, regional variations in colloquial Bangla, develop to distinct linguistic nuances and contextual complexities that render accurate translation challenging (Bentivogli et al, 2016)\cite{bentivogli2016neural}. In recent times, digitalization and globalization have made the availability of a Bangla MT an unavoidable necessity for global communication, especially as access to digital devices has become widespread and even for monolingual users in rural areas. When these users attempt to learn or access information online, existing MT systems often fail to produce natural translations from colloquial input, often leading to poor or ambiguous results. Developing the capacity to use informal language sets is therefore crucial to successfully assisting millions of Bangla users within the digital domain. \\[5pt]\

In order to meet this challenge, the research in this article focuses on the creation of an open source informal language set and the development of appropriate machine translation models specialized in the Bangla informal language.

\subsection{Research Objectives}
This study systematically investigates the challenges inherent in informal, low-resource Bangla machine translation. To achieve this, the primary objectives of the work are as follows:
\begin{itemize}
  \item To investigate the main barriers encountered in low-resource Bangla machine translation.
  \item To evaluate state-of-the-art MT approaches for Bangla to identify critical gaps requiring improvement in informal language translation.
  \item To construct an informal Bangla to English training dataset to resolve the lack of available data and improve translation capabilities.
  \item To research the characteristics of informal Bangla language and provide an analysis comparing the predicted outputs of different models.
  \item To explore data augmentation strategies specifically targeted at the informal datasets to successfully enhance overall data size.

\end{itemize}

%% file: chapters/chapter_2.tex
Zhang et al. (2020) \cite{zhang2020improving} aimed to improve multilingual and zero-shot translation, where models often struggle, especially with many languages or unseen pairs. They used a deeper Transformer encoder–decoder, hypothesizing that limited capacity and lack of parallel data cause errors. Artificial parallel data was generated for zero-shot pairs. Training data came from OPUS, capped at 1M pairs per language, with 2,000 reserved for validation and testing. Cross-lingual filtering avoided overlap, resulting in the OPUS-100 dataset (55M pairs). For zero-shot evaluation, 2K test pairs were sampled for 15 language pairings. They evaluated one-to-many and many-to-many scenarios using byte-pair encoding (64k vocab), testing configurations including baseline Transformer, Language-Aware Layer Normalization (LALN), Language-Aware-Linear-Transformation (LALT), and their combinations; in many-to-many mode, Random Online Back-Translation (ROBT) was also used. Merged attention mechanisms reduced training time. The combination of Transformer, LALN, LALT, and merged attention achieved the highest win ratio (92.6\% one-to-many, strong in many-to-many), while ROBT slightly reduced this ratio. \\[5pt]

Wang et al. (2020) \cite{wang2020balancing} introduced a technique using an optimized data scorer to weight training data, maximizing performance across languages. Differentiable Data Selection uses bi-level optimization to dynamically adjust weights based on validation improvements. To maintain scalability and generality, they extended it to MultiDDS. The approach was tested with a standard Transformer on parallel data from 58 languages to English, in two settings: Related (four low-resource languages each linked to four high-resource ones) and Diverse (eight unrelated languages of varying sizes). Evaluations covered many-to-one and one-to-many tasks, measured with BLEU. MultiDDS outperformed temperature-based and proportional-based heuristics in both settings, with the best results from the "High" optimization attribute, which prioritizes high-performing languages. \\[5pt]

Bhattacharjee et al. (2021) \cite{bhattacharjee2021banglabert}, researched to improve Natural Language Understanding (NLU) for low-resource languages like Bangla, that actually perform poorly on NLU tasks. So, they came with BanglaBERT, a BERT model pre-trained on Bangla data, and BanglishBERT that was trained on both Bangla and English. Due to limited Bangla resources online, they went through 110 websites and ranked them by Amazon Alexa, as well as collected contents from encyclopedias, news, blogs, e-books, stories, and social media. The raw data set was 35 GB and after preprocessing steps such as deduplication, removal of HTML and JavaScript tags, and filtering non-Bangla pages using a language classifier, the final dataset was 27.5 GB, containing 5.25M documents with an average of 306.66 words each. BanglaBERT was trained using ELECTRA and Adam, while BanglishBERT used a balanced English–Bangla corpus. They introduced the Bangla Language Understanding Benchmark (BLUB), covering sentiment classification, natural language inference, named entity recognition, and question answering. BanglaBERT achieved the highest BLUB score (77.09) under supervised fine-tuning, while BanglishBERT (75.73) and other models showed lower adoption. \\[5pt]\

In an effort to simplify NMT architectures, \cite{you2020hard}, You et al. (2020) find out a hard-coded Gaussian attention mechanism that excludes all learned parameters from the attention process. Traditionally, Transformer models rely heavily on multi-headed attention, which is sometimes fast to train and also confusing on its actual impact on translation quality. So, they replaced attention heads with Gaussian distributions centered at specific positions of input sequence. This ideology led to comparable BLEU scores with four language pairs, while improving memory efficiency by 26.4\% per batch and boosting decoding speed by 30.2\% per second. Studies began with IWSLT16 English–German and extended to Romanian, French, and Japanese. When both self-attention and cross-attention were hard-coded, translation quality dropped, showing cross-attention's importance. The decoder's limitations likely contributed to poorer performance, while feedforward layers functioned well. Longer sentences were challenging, especially when subject-verb relationships were distant. Their work lays the groundwork for improved attention techniques in NMT. \\[5pt]\

According to Baziotis et al. (2020) \cite{baziotis2020language}, explored the limitations caused by lack of large-scale parallel corpora in NMT, especially for low-resource languages. They proposed using a target-side monolingual language model as a prior, combined with posterior regularization, to guide translation and improve accuracy. The method uses minimal monolingual data and leads to faster decoding than previous approaches relying on back-translation or noisy channel models. Positive outcomes were seen for English–German and English–Turkish. After data augmentation, 3 million texts were used for Turkish. Both models were trained with Adam (5000 tokens per batch, 10 iterations). The LM-prior led to a 1.8 BLEU score improvement. English–German translation performed well with about 30 million German sentences. The LM-prior gradually performs better in all directions, providing a scalable solution for low-resource languages. \\[5pt]\

Edunov et al. (2020) \cite{edunov2019evaluation} studied Back Translation (BT) in NMT, finding that human translators preferred BT outputs for their fluency and naturalness. BT was tested on English–German, German–English, English–Russian, and Russian–English, using WMT '18 and '19 news data. Both parallel and back-translated data were trained with the Facebook FAIR system and Fairseq tools. Original and back-translated outputs showed little difference, and BT improved training efficiency. The study addressed BLEU's limitations and evaluated BT through human fluency surveys, with 37\% support in English–Russian and 28\% in German–English. For future research, the authors recommend integrating BLEU with language model scores. \\[5pt]\

In order to enhance representation learning, Yin et al. (2020) \cite{yin2020novel},  presented a graph-based multi-modal fusion encoder for NMT, fusing text and visual features. The encoder creates a unified network from input sentences and images, using semantic exchange to enrich node representations. Multiple fusion layers further refine these representations, providing the decoder with an attention-based context vector. The model outperformed most current approaches on Multi30K dataset for English–German translation and matched METEOR scores of Trg mul RNN and Fusion conv RNN on WMT2017. Their work highlights the value of explicit cross-modal semantic modeling and suggests a more effective approach for multi-modal NMT. \\[5pt]\

Bugliarello and Okazaki (2020) \cite{bugliarello2019enhancing}, presented Pascal, a modified self-attention technique that incorporates syntactic structure into the transformer model to increase translation accuracy. They observe that most neural translation systems depend on parallel phrase pairings and assume that attention layers implicitly learn syntax, an assumption that breaks down for long sentences and low resource languages. Pascal operates as a local attention mechanism with no extra trainable parameters, adjusting token scores according to their distance from the position of their syntactic parent. Hanse, it enhances model understanding while that is simple by incorporating parent context into word representations directly. Otherwise,  Pascal consistently outperformed the baseline transformer in the English-German, English-Turkish, and English-Japanese translations that were reviewed by using data from the WMT and WAT datasets. In particular, Pascal delivered up to 0.9 BLEU points and 1.75 RIBES improvements compared to previous syntax integration techniques. Despite these promising results, evaluation remains limited to a small set of language pairs and there is no direct comparison with other state-of-the-art models. All things considered, this research illustrates the advantages of precise syntactic attention and offers unique ideas for effective syntax modeling in machine translation. \\[5pt]\

The architecture No Language Left Behind (NLLB-200), consisting of an encoder and decoder that uses transformers, was created by NLLB Team et al. (2022) \cite{NLLB2022}. It is able to provide good translation quality for 200 language pairs, even the less frequently translated ones and balances the data and addresses the training specific language issues. This research uses language adaptive modules, and in order to cope with the challenges of scaling, the team uses a variety of filtering, balanced language data, and improved tokenization with SentencePiece. It is common knowledge that the architecture is over-fitted in multilingual datasets to improve cross-lingual transfer. These various methods and the overall system design to provide translation in less represented languages is an excellent example and a milestone for the rest of the systems. \\[5pt]\

Lin et al. (2020) \cite{lin2020pre} introduced mRASP, a universal multilingual NMT technique pre-trained with 32 bilingual datasets using random aligned substitution (RAS). RAS brings semantically similar words and phrases across languages closer in representation space. mRASP was tested on 42 translation directions in low, medium, and high-resource scenarios, showing significant gains over direct training. Zero-shot translation was evaluated in four scenarios: Exotic Pair, Exotic Source, Exotic Target, and Exotic Full. Fine-tuning on Exotic Full still outperformed direct training. Unlike other methods, mRASP achieves strong results with only a few million pairs. The paper stresses the need for aligned objectives between pre-training and fine-tuning and suggests future work on alternative alignments and larger datasets. \\[5pt]\

Wang et al. (2020) \cite{wang2020multi}, proposed the multitask learning system that improves Multilingual Neural Machine Translation (MNMT) performance by utilizing bitext and monolingual data. They observe that monolingual resources remain underused in MNMT, so they combine standard translation tasks on parallel corpora with denoising objectives on monolingual text. For both high and low resource languages, the method improved performance on 10 language pairs from WMT datasets, especially for zero-shot settings. Translation quality was further improved with dynamic temperature-based sampling and dynamic noising ratio strategies. The research was limited to a small number of language pairs and did not explore alternative denoising techniques or multitask architectures. The authors suggest benchmarking against recent MNMT architectures, increasing language coverage, and enhancing resource efficiency to address these gaps. \\[5pt]\

Islam et al. (2021) \cite{islam2010english},  proposed a phrase-based statistical machine translation system to convert English to Bangla by adding transliteration and preposition handling modules to a baseline. While the preposition handler modifies word order by moving post-positional words before their reference objects during preprocessing and reversing this step with suffix addition in postprocessing, the transliteration module converts out-of-vocabulary terms into Bangla script to remove any remaining English words in the output. BLEU, NIST, and TER were used to evaluate the system, which was dependent on parallel corpora for training. Against the open-source Anubadok system, which scored 1.60 BLEU, their enhanced model achieved 11.70 BLEU, confirming the utility of the new modules. The authors observe that the system excels in short sentence but may still retain untranslated English words. They suggested that future work should expand on parallel data, introduce varied test sets with multiple references and integrate extra post-positional words, inflectional suffixes, compound word handling and other linguistic features to further boost translation quality. \\[5pt]\

In 2023 Zhao and others \cite{Zhao2023QLoRA} published QLoRA, a method for scaling the fine-tuning of large language models for the tasks of quantization and low-rank adaptation. This method of scaling loads the models in 4-bit NF4 quantization, yielding a drastic reduction in the memory footprint of the models and training LoRA modules in training on the query, key, value, and output projections for the query and value layers. This approach allows a limited fraction of parameters to be updated for the LoRA modules, thus facilitating training and maintaining training performance. QLoRA has been shown in several tests to produce results equal to those of fully fine-tuned models, but at a fraction of the computational cost of such models. The work shown to enable the adaptation of cutting-edge models sits at the intersection of low resource availability and the monolingual and domain translation tasks.

%% file: chapters/chapter_3.tex
The development of high-performing NMT systems is mainly limited to large scale high quality parallel corpora leading to issues for low-resourced languages such as Bangla. Although there are large multi-domain Bengali MT datasets such as the BanglaNMT, which contain millions of pairs, they are of little use because they are mostly composed of automatically converted texts or it is heavily biased towards formal domains (news, books, legislation, etc.). For the specific purposes of our research, it is necessary to have a parallel corpus containing informal, colloquial and conversational Bengali, as these sources are very little available.
\renewcommand{\thefigure}{3.\arabic{figure}}
\setcounter{figure}{0}
\subsection{Dataset Construction and Characteristics}
Our dataset is a unique, custom-built, translative collection composed of 7,664 pairwise Bangla-to-English sentences. This dataset was compiled to capture the non-standard linguistic features of informal communication prevalent in digital media. \\[5pt]\

Data was predominantly gathered from social media sites, including comments, posts, and statuses on Facebook and YouTube, augmented by typical informal conversation sentences. The collection targets the problem of informal texts- code switching, slang, and dialectal variations that are prolific in user-generated content. \\[5pt]\

The raw Bangla sentences were manually translated by native Bangla and bilingual English speakers to produce the equivalent English output sentences. This labor-intensive procedure was important to ensuring semantic accuracy and capturing the intended vernacular meaning, which automated translation methods sometimes fail to do.
The typical samples in Table \ref{tab:bn-en-examples} demonstrate the linguistic variation and colloquial patterns of our corpus.

\vspace{0.2 cm}
\begin{table}[htbp]
  \caption{Examples of Informal Bangla-English Translations}
  \label{tab:bn-en-examples}
  \centering
  \renewcommand{\arraystretch}{1.2} 
  \begin{tabularx}{\textwidth}{@{} p{0.5\textwidth} X @{}} 
    \toprule
    \textbf{Bangla} & \textbf{English} \\
    \midrule
    {\bengalifont ভেবে করিও কাজ, করিয়া ভেবো না।} & Think before doing, don’t think after doing. \\
    \midrule
    {\bengalifont বাঙালি পারলে মেট্রোর ছাদে উঠে যাইত খালি এই মেট্রো গুলা কারেন্টে চলে দেখে পারতাছে না।} & Bengali would travel on top of the metro, but they couldn’t as these run on electricity. \\
    \midrule
    {\bengalifont আতিফ ইসলাম আসুক আর যেই আসুক আমার আম্মু আমাকে যাইতে দিবেনা।} & Atif Islam or whoever comes, my mother will not let me go. \\
    \midrule
    {\bengalifont মামা টাকা কথা কয় ভুলে গেছো।} & Mama, have you forgotten that money talks? \\
    \midrule
    {\bengalifont সাবাস, দেরিতে হলেও বাংলার মানুষ বুঝতে শিখেছে।} & Bravo, people of Bengal have started understanding even though it is late. \\
    \midrule
    {\bengalifont কেন ভাই, বাংলাদেশে কি উট পালন করা যায় না?} & Why brother, can camels not be raised in Bangladesh? \\
    \bottomrule
  \end{tabularx}
\end{table}

\subsection{Analysis of Dataset}

The initial dataset of 7,664 pairs contained a total of 65,754 Bangla words and 85,402 English words. An analysis of the sentence lengths for both the Bangla and English texts, categorized by word count (short: 1–5; medium: 6–10; long: 11–15; very long: >15), revealed a distinct, positively skewed distribution towards shorter sentences. \\[5pt]\

\begin{figure*}[htbp]
  \centering
  \begin{subfigure}[t]{0.48\textwidth}
    \centering
    \includegraphics[width=\linewidth]{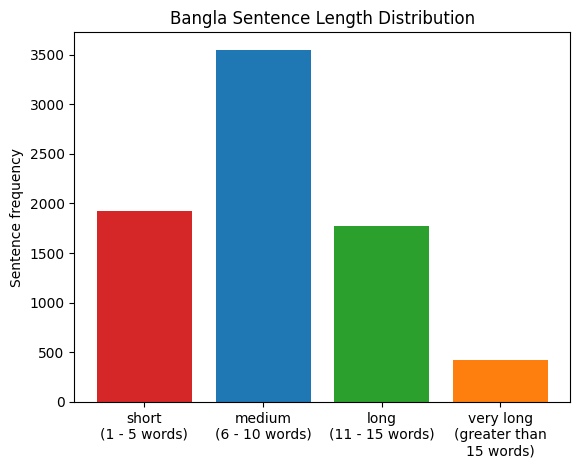}
    \caption{Bangla Sentence Length Distribution}
    \label{fig:x-bangla-length}
  \end{subfigure}
  \hfill
  \begin{subfigure}[t]{0.48\textwidth}
    \centering
    \includegraphics[width=\linewidth]{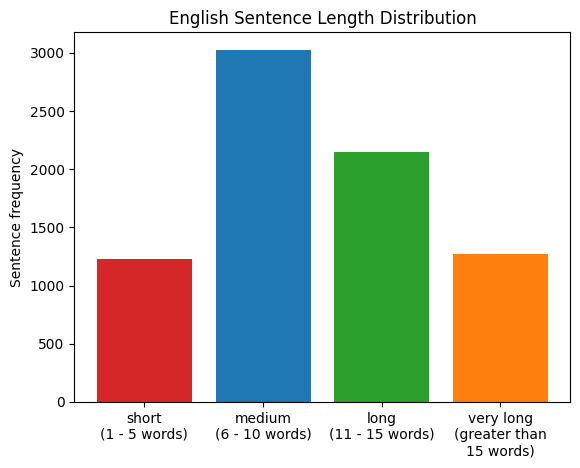}
    \caption{English Sentence Length Distribution}
    \label{fig:x-english-length}
  \end{subfigure}
  \caption{Sentence length distributions for Bangla and English datasets.}
  \label{fig:x-lengths}
\end{figure*}

In the Bangla portion, the majority of content consisted of medium sentences (3,548), followed by short (1,927), long (1,768), and finally very long (421) sentences. In contrast, the English side comprised 3,025 medium sentences, 2,145 long sentences, 1,268 very long sentences, and 1,226 short sentences. This distribution, visualized in Figure \ref{fig:x-bangla-length} and \ref{fig:x-english-length}, confirms the conversational and concise nature of the collected informal discourse. \\[5pt]\

This distribution confirms the conversational and concise nature of the collected informal discourse. The most frequently occurring non-stop-word Bangla words were {\bengalifont 'একটা'} (one, 355 occurrences), {\bengalifont 'ভাই'} (brother, $\approx$300), and {\bengalifont 'মানুষ'} (people, $\approx$250). \\[5pt]\

We can observe the length distribution of both the languages from Figure \ref{fig:x-bangla-length} and \ref{fig:x-english-length}. A sentence length of 1 to 5 is classified as a short sentence, 6 to 10 words is classified as a medium sentence, 11 to 15 words is classified as a long sentence, and a sentence length of greater than 15 words is classified as a very long sentence.

As can be observed from the figures below, the dataset follows a positively skewed distribution for both Bangla and English languages in terms of sentence length.

\vspace{0.2cm}

\begin{figure}[htbp]
  \centering
  \begin{subfigure}[t]{0.48\textwidth}
    \centering
    \includegraphics[width=\linewidth]{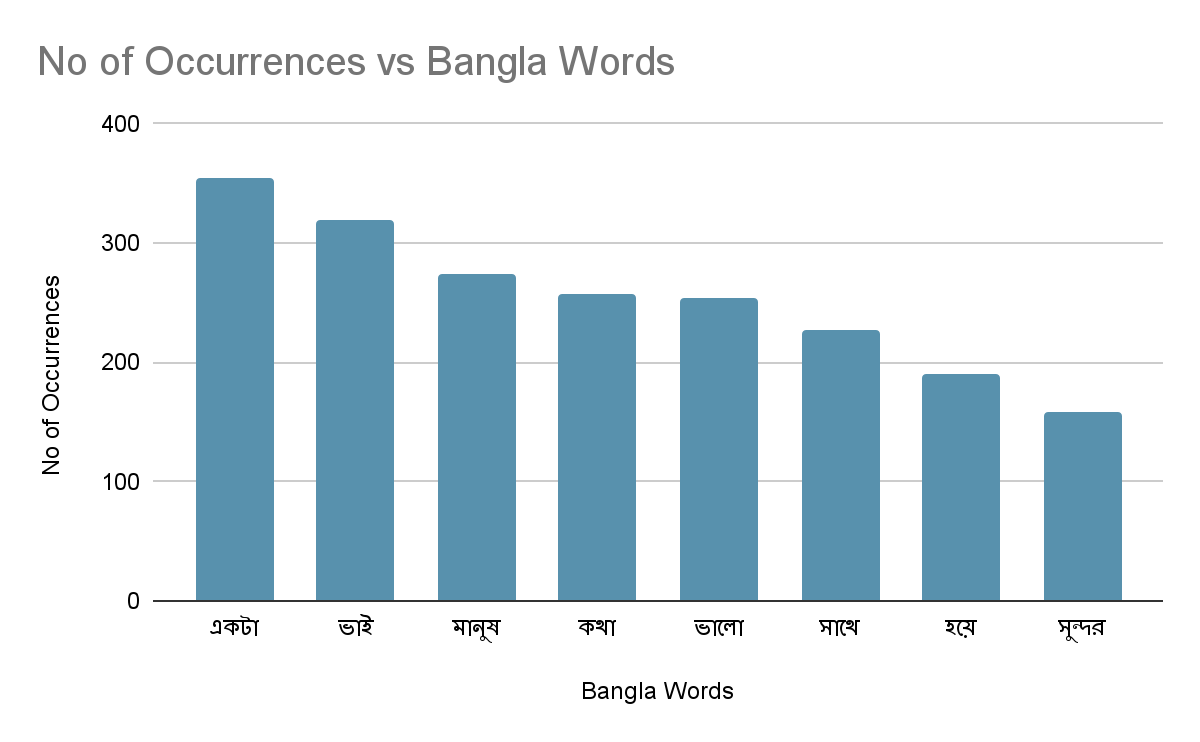}
    \caption{Most common Bangla words occurrence}
    \label{fig:x-Bangla-Word-Occurrence}
  \end{subfigure}
  \hfill
  \begin{subfigure}[t]{0.48\textwidth}
    \centering
    \includegraphics[width=\linewidth]{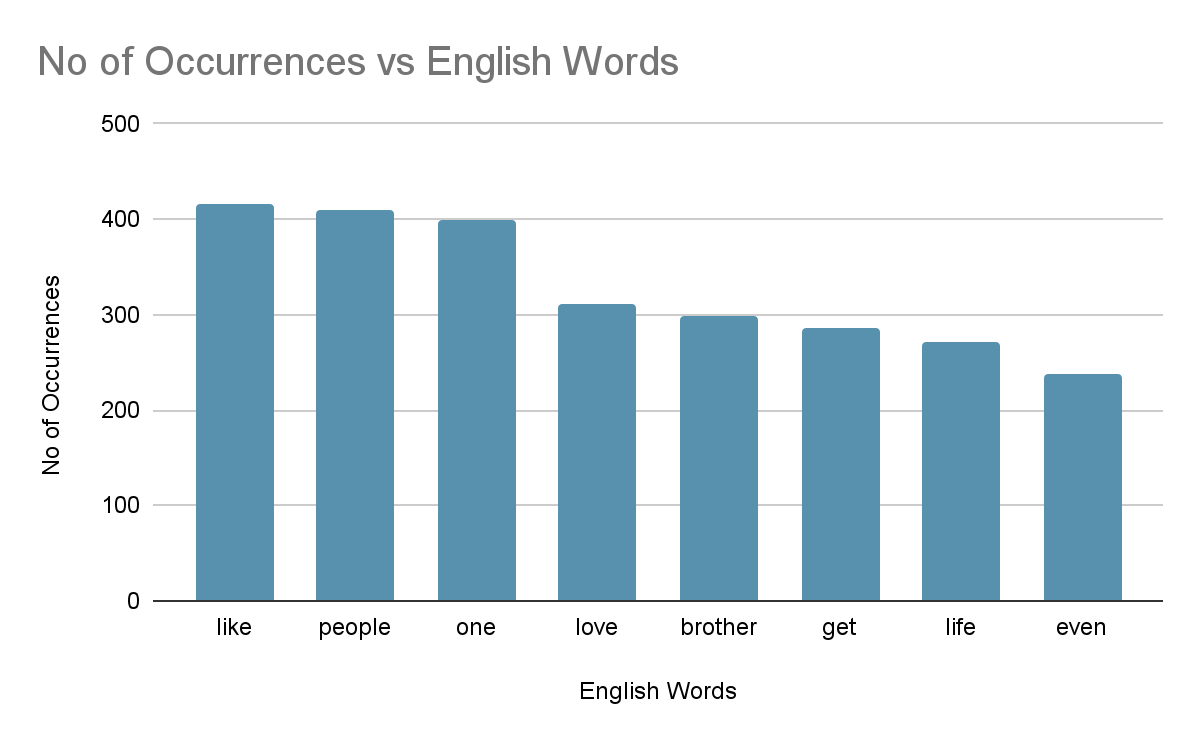}
    \caption{Most common English words occurrence}
    \label{fig:x-English-Word-Occurrence}
  \end{subfigure}
  \caption{Comparison of the most common Bangla and English word occurrences.}
  \label{fig:x-Word-Occurrences}
\end{figure}

In Figure \ref{fig:x-Bangla-Word-Occurrence}, {\bengalifont "একটা"} is observed to be the most common Bangla word, with a frequency of 355, after excluding formal Bangla stopwords. The second word in the rank is a little above 300, after which {\bengalifont "ভাই", "মানুষ", "কথা"}, and {\bengalifont "ভালো"} have a frequency of around 250. The next two, {\bengalifont "হয়ে"} and {\bengalifont "সুন্দর"}, are below 200, and as the plot shows, each appears less than 200 times. The adjacent figure \ref{fig:x-English-Word-Occurrence} similarly depicts most used English words, with "like" being the most common, at 415 instances, after excluding English stop-words. The next two, "people" and "one", are each around 400. Meanwhile, "love", "brother", "get", and "live" are each over 200, averaging around 250, and "even" is just below 250, as stated in the previous sentence.

\subsection{Data Augmentation}
To achieve adequate performance in NMT models, particularly with architectures possessing large parameter counts, a significantly larger dataset is typically required. As Bangla is a low-resource language and the manual data collection process is resource-intensive, two data augmentation strategies were employed to expand the dataset size: \textbf{Back-translation (BT)} and \textbf{Synonym Replacement}. This process is consistent with strategies used in other low-resource MT studies to improve model generalization and robustness.

\subsubsection{Back-translation}
In this procedure, the English output sentences were manually translated back into a semantically equivalent, but stylistically varied, Bangla input sentence. This generated a new parallel pair that captured similar informal context. For example:

\begin{center}
Original Bangla sentence: {\bengalifont এক্কেবারে হাছা কথা কইছেন।}
\end{center}
\begin{center}
English Sentence: You have said the truth
\end{center}
\begin{center}
Back-translated to Bangla: {\bengalifont একদম সত্যি কথা বলেছেন।}
\end{center}
\vspace{0.5cm}
\subsubsection{Synonym Replacement}
The original Bangla text underwent systematic substitution of words with their most common informal synonyms to create orthographic and lexical diversity while preserving the intended meaning. As most of the words are informal, we used Facebook's pretrained FastText \footnote{\href{https://fasttext.cc/docs/en/crawl-vectors.html}{https://fasttext.cc/docs/en/crawl-vectors.html}} embedding to get the closest representation of those words. Manual intervention was required for cases where automated lookup was not possible. For example:

\begin{center}
Original Bangla sentence: {\bengalifont মামা, টাকা কথা কয় ভুলে গেছো।}
\end{center}
\begin{center}
English Sentence: Mama, have you forgotten that money talks
\end{center}
\begin{center}
Back-translated to Bangla: {\bengalifont মামা, টাকা কথা বলে ভুলে গেছো।} 
\end{center}
\vspace{0.2cm}
Here, the Bangla word {\bengalifont "কয়"} is replaced with {\bengalifont "বলে"}, which holds similar meaning.
\vspace{0.5cm}
Overall, this study performed data augmentation on \textbf{7003} samples of the data set. As a result, our final data set comprised \textbf{14,667} sets of Bangla and English sentences.

\vspace{0.5cm}
\subsection{Data Preprocessing}
The dataset was subjected to minimal but essential preprocessing steps to ensure data quality and compatibility with the NMT models.
\subsubsection{Data Cleaning}
As the data set was manually collected and translated, researchers do not have to follow strict preprocessing. For both Bangla and English sentences, non-alphanumeric characters, special characters, punctuation and extra white spaces were removed. Additionally, the texts were converted to lowercase for the English sentences. Furthermore, modification performed on the English contractions e.g. \textbf{I'm} to \textbf{I am}.

\textbf{Before Cleaning:}
\begin{center}
    Bangla: {\bengalifont ভেবে করিও কাজ, করিয়া ভেবো না।}
\end{center}
\begin{center}
    English: Think before doing, don't think after doing.
\end{center}

\vspace{0.5cm}

\textbf{After Cleaning:}
\begin{center}
    Bangla: {\bengalifont ভেবে করিও কাজ করিয়া ভেবো না}
\end{center}
\begin{center}
    English: think before doing do not think after doing
\end{center}

\subsubsection{Data Splitting}
Dataset were split into 80:10:10. 80 percent for training set, 10 percent for validation set, and 10 percent for testing set. So 11,733 sentences used for train each models, and 1,467 sentences used for each Validation and Test
\begin{itemize}
    \item \textbf{Train:} 11,733 sentences
    \item \textbf{Validation:} 1,467 sentences
    \item \textbf{Test:} 1,467 sentences
\end{itemize}

Due to the small dataset, researchers had to keep the training set adequate while still maintaining a viable validation and testing set. Even though, losing the size of the training set, it is important to have the validation set. This is due to the requirement of early stopping, in order to prevent over-fitting of the model. \\[5pt]\

Overall, the dataset has \textbf{7,664} (collected and translated) + \textbf{7,003} (augmented and translated) = \textbf{14,667} pairwise sentences. Ultimately, the training dataset size is \textbf{11,733} which would be used for training three models: \textbf{BiLSTM}, \textbf{mt5-small} and \textbf{mt5-large}. While training, the other \textbf{1,467} samples would be used for validating. Finally, the remaining \textbf{1,467} samples would be used for testing.

%% file: chapters/chapter_4.tex
A comparative evaluation of three distinct modeling architectures was conducted in order to tackle the difficulties associated with informal, low-resource Bangla machine translation. The first system uses a Bi-Directional Long Short-Term Memory (BiLSTM) with an attention mechanism, which is a type of recurrent neural network (RNN), serving as a primary baseline due to the integration of sequence-to-sequence dynamics (Cho et al, 2014; Bahdanau et al, 2014)\cite{cho2014learning}\cite{bahdanau2014neural}. The second approach refines two versions of the Multilingual Text-to-Text Transfer Transformer (mT5) which is built on the latest transfer learning framework (Raffel et al, 2020; Xue et al, 2021)\cite{raffel2020t5}\cite{xue2021mt5}. Finally, to push the state-of-the-art, the third approach leverages the \textbf{No Language Left Behind (NLLB-200)} model, a massive multilingual transformer explicitly optimized for low-resource languages (NLLB Team et al, 2022)\cite{NLLB2022}. The final, expanded dataset of 14,667 sentence pairs, as detailed in Section 3, was used to train and assess all models.

\subsection{BiLSTM with Attention Mechanism}
The initial system to translate, includes an attention mechanism was added to the Bidirectional LSTM (BiLSTM) Encoder-Decoder architecture, contains two LSTM layers: one for processing input in the forward direction, and the other one is for processing in the backward direction. This architecture was selected due to RNNs' intrinsic ability to analyze sequential inputs, and the bidirectional aspect enables the encoder to record context from both prior and following words, an essential characteristic for managing the variable word order prevalent in Bangla and informal language (Popović, 2017)\cite{popovic2017comparing}.

\subsubsection{Word Embedding}
Standard word embeddings like GloVe frequently fail to represent the colloquialisms and out-of-vocabulary (OOV) terminology found in informal Bangla writings from social media.  To mitigate this problem, a representation of the input text is done using \footnote{\texttt{https://fasttext.cc/docs/en/crawl-vectors.html}} Facebook’s pre-trained FastText library. (Bhattacharjee et al., 2021)\cite{bhattacharjee2021banglabert}. The entire 300-dimensional vector \textbf{(cc.bn.300.bin)} of the word has been loaded and the dimension has been computed to \textbf{100} for faster convergence in training. This approach has made it possible to map even highly informal or misspelled tokens into meaningful vector representations.

\subsubsection{Encoder-Decoder Architecture}
The BiLSTM architecture follows a classic sequence-to-sequence design, visually represented in Figure \ref{fig:x_bilstm}. \\[5pt]\

The \textbf{Encoder Layer} takes tokenized Bangla sequence as input. This passes through the Embedding Layer, where the research finds to set the dimension to 100. A \textbf{BiLSTM layer}, configured to have a latent dimension of \textbf{128}. This design results in \textbf{256} effective LSTM units, allowing the encoder to thoroughly capture both forward and backward contextual dependencies within the source sentence. \\[5pt]\

Next, the system integrates an \textbf{Attention Mechanism}. The encoded states are directed through an additive attention layer. This layer is essential for dynamically focusing the decoder's attention, employing a repetition vector and a concatenation layer prior to processing via two dense layers for scoring and normalization via the \textbf{Softmax activation}. The purpose behind including attention layer is, it significantly improves translation quality, especially when working with resource-constrained, low-resource settings, by enabling the decoder to pinpoint the most relevant source segments for generating each target token (Bahdanau et al, 2014)\cite{bahdanau2014neural}. \\[5pt]\

The \textbf{Decoder Layer} is a unidirectional LSTM comprising \textbf{256} units. It processes the context vector received from the attention mechanism. The output of the decoder is then mapped to the target English vocabulary by the output layer, which is a dense layer. This technique utilizes a softmax activation function to generate a probabilistic distribution for the subsequent anticipated word. \\[5pt]\

The model was trained by minimizing the \textbf{Categorical Cross-Entropy Loss} ($\mathcal{L}$), which is standard for sequence-to-sequence tasks like the study.
\[
\mathcal{L} = - \frac{1}{N} \sum_{i=1}^{N} \sum_{t=1}^{T} \log(\hat{y}_{i,t} \cdot y_{i,t})
\]
For model optimization, the study uses \textbf{Adam optimizer} and set the starting learning rate at 0.001.

\renewcommand{\thefigure}{4.\arabic{figure}}
\setcounter{figure}{0}

\begin{figure}[H]
    \centering
    \includegraphics[width=0.8\textwidth]{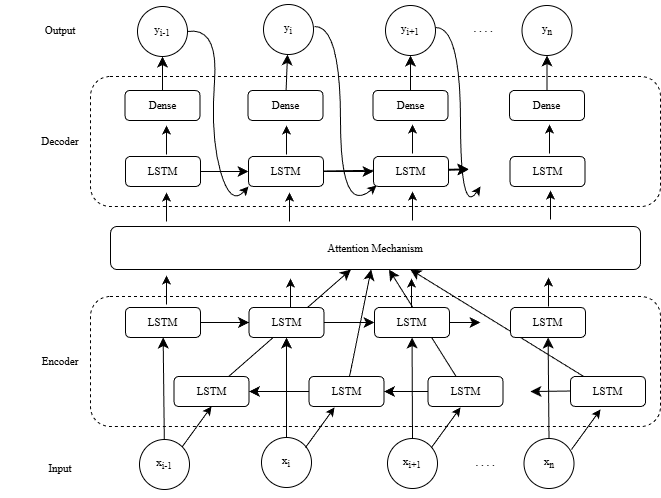}
    \caption{Encoder-Decoder Architecture}
    \label{fig:x_bilstm}
\end{figure}

\subsection{mT5}
The text-to-text transformer (T5) (Raffel et al., 2020) \cite{raffel2020t5} model has been pre-trained on a large number of corpora texts. It is trained for tasks including text categorization and language modeling. Transfer learning may be applied to downstream tasks such machine translation, question-answering, and sentiment analysis. Furthermore, transfer learning will enhance the efficacy of low-resource language models. Consequently, the research will be refining the model using our dataset, which is a sort of transfer learning. \\[5pt]\

T5 consists of encoder and decoder, where there are multiple encoder layers and decoder layers, respectively. Each layer also has multiple attention heads that are responsible for working with different parts of the input sequence. It also follows masked language modeling. This is the process of hiding some words in a sequence randomly with tokens, so that the model can predict those words that were hidden. \\[5pt]\

We used Multilingual T5 (mT5) (Xue et al, 2021)\cite{xue2021mt5} which has a similar architecture to T5 and has been pre-trained on the Common Crawl-based dataset which involves 101 languages. As Bangla is also part of these languages, we conduct our research mT5 for our translation task. However, unlike T5, mT5 had unsupervised pre-training and hence, it needs to be fine-tuned before it can work for machine translation.

\begin{figure}[H]
  \centering
  \includegraphics[width=0.9\linewidth]{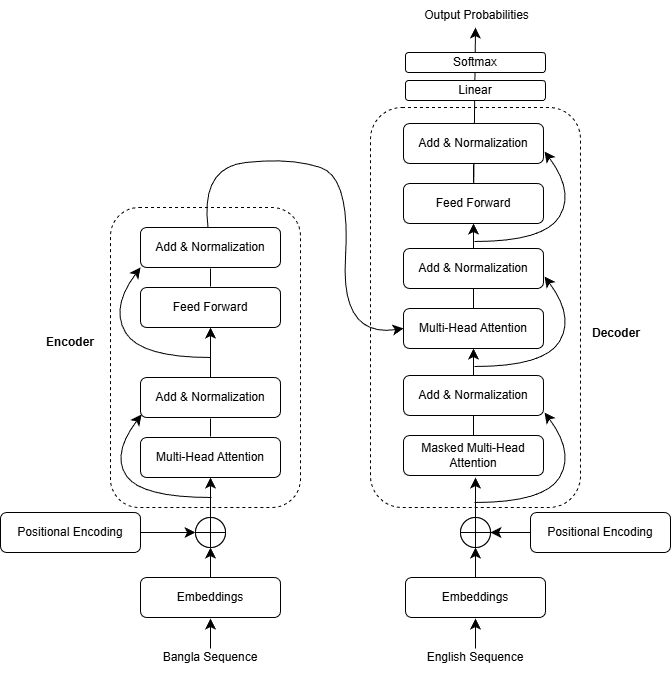}
  \caption{\centering mT5 Architecture}
  \label{fig:mt5-architecture}
\end{figure}

\subsubsection{Tokenizer}
For Bangla text tokenization, we used a \footnote{\href{https://huggingface.co/csebuetnlp/banglat5}{https://huggingface.co/csebuetnlp/banglat5}}pre-trained Bangla tokenizer. For instance:

\begin{center}
Bangla Sentence: \texttt{{\bengalifont ভাই একটু কম পানিতে যান সর্দি লেগে যাবে}}
\end{center}

\begin{center}
Tokens: \texttt{[{\bengalifont'\_ভাই', '\_একটু', '\_কম', '\_পানিতে', '\_যান', '\_সর্দি', '\_লেগে', '\_যাবে',} '</s>']}
\end{center}

\begin{center}
Corresponding IDs: \texttt{[  290,   123,   276,  3078,   408, 16969,  1576,   154,     1]}
\end{center}

For English text tokenization, we used the default sentence piece tokenizer, which is provided by mT5.

\begin{center}
English Sentence: \texttt{brother please go into the water a bit less you will catch a cold}
\end{center}

\begin{center}
Tokens: \texttt{['\_brother', '\_please', '\_go', '\_into', '\_the', '\_water', '\_', 'a', '\_bit', '\_less', '\_you', '\_will', '\_', 'catch', '\_', 'a', '\_cold', '</s>']}
\end{center}

\begin{center}
Corresponding IDs: \texttt{[58276, 10151,  1002,  2387,   287,  4582,   259,   262,  5485, 24691,
           521,   898,   259, 15249,   259,   262, 44271,     1]}
\end{center}

\subsubsection{mT5-small and mT5-large}
This study included the fine-tuned two mT5 pretrained checkpoints in our data set; \textbf{mT5-small} and \textbf{mT5-large}. The mT5-small has \textbf{300 million parameters} and mT5-large has \textbf{1.2 billion}. Since the dataset is small, this work initially fine-tuned mT5-small due to its smaller parameters. However, due to mediocre results, the study eventually move to mT5-large. \\[5pt]\

We used the \textbf{default configurations} for both mT5-small and mT5-large. The checkpoints were trained for \textbf{10 epochs} with \textbf{batch size = 8}. We set the \textbf{initial learning rate} as \textbf{$3 \times 10^{-4}$} and \textbf{Adam} as the \textbf{optimizer}.

\subsection{NLLB-200 Transformer (QLoRA Fine-Tuning)}

To benchmark state-of-the-art performance and fully harness the capabilities of massively multilingual models, we extended our research to incorporate the NLLB-200 (No Language Left Behind) architecture (NLLB Team et al., 2022) \cite{NLLB2022}. NLLB-200 is an advanced Transformer-based encoder-decoder model explicitly designed to deliver high-quality translation, including for low-resource languages, by supporting 200 language pairs. 
This experiment evaluated three NLLB-200 checkpoint variants to investigate scalability and translation quality: the distilled 600 million parameter model, the 1.3 billion parameter model, and the larger 3.3 billion parameter model.

\subsubsection{Tokenizer}

NLLB-200 employs a SentencePiece tokenizer similar in design to mT5, but with a critical distinction: \textbf{the use of explicit language codes}. Each input and output sequence must be prefixed with a language identifier, which guides the model during both training and inference.

For Bangla--English translation, the following language tags were used throughout this study:
\begin{itemize}
    \item \texttt{ben\_Beng}: Bangla (Bengali script)
    \item \texttt{eng\_Latn}: English (Latin script)
\end{itemize}

Unlike mT5, which implicitly learns translation direction, NLLB requires explicit prefixing. Input Bangla sentences were formatted as:
\[
\texttt{<ben\_Beng>} \quad x_{1}, x_{2}, \ldots, x_{T}
\]

Similarly, target English sequences were constructed as:
\[
\texttt{<eng\_Latn>} \quad y_{1}, y_{2}, \ldots, y_{T}
\]

Both source and target sequences were truncated or padded to a maximum of \textbf{256 tokens}, balancing computational efficiency and context retention.

\begin{figure}[H]
  \centering
  \includegraphics[width=0.9\linewidth]{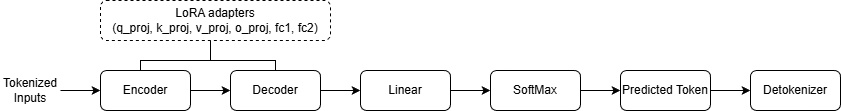}
  \caption{\centering NLLB+LoRA Flowchart}
  \label{fig:NLLB+LoRa_flowchart}
\end{figure}
\subsubsection{NLLB Variants and QLoRA Fine-Tuning}

Direct full-parameter fine-tuning of the 1.3B and 3.3B parameter NLLB models is computationally prohibitive on consumer GPUs. Therefore, this study adopts \textbf{Quantized Low-Rank Adaptation (QLoRA)} (Zhao et al., 2023; Hu et al., 2021)\cite{Zhao2023QLoRA}\cite{Hu2021LoRA}, a highly efficient parameter-efficient fine-tuning (PEFT) method.

\paragraph{4-bit Quantization (NF4).}
The base model weights were loaded in \textbf{4-bit NF4 quantization}, reducing GPU memory usage by more than 60\%, enabling large models to be fine-tuned on a single RTX 4070 Super GPU.

\paragraph{LoRA Module Injection.}
Trainable low-rank adaptation matrices were inserted into key Transformer components, namely the \texttt{q\_proj}, \texttt{k\_proj}, \texttt{v\_proj}, \texttt{o\_proj} projections and the \texttt{fc1}/\texttt{fc2} feed-forward sublayers. This setup updates only \textbf{2--3\% of the full model parameters}, while the remaining weights remain frozen, ensuring efficient yet expressive fine-tuning.

\paragraph{Training Configuration.}
Across all three NLLB variants, the following hyperparameters were used:
\begin{itemize}
    \item \textbf{Epochs}: 10
    \item \textbf{Learning Rate}: $2 \times 10^{-5}$
    \item \textbf{Optimizer}: AdamW
    \item \textbf{Precision}: Mixed FP16 training
    \item \textbf{Loss Function}: Cross-entropy loss
\end{itemize}

The design prioritizes computational efficiency without sacrificing translation performance.

\subsubsection{Beam Search and Oracle Rescoring}

During inference, the system utilizes \textbf{Beam Search} with a beam width of 5 (Koehn \& Monz, 2006)\cite{koehn2006manual}, generating five translation candidates for each Bangla sentence:
\[
\hat{Y} = \{y^{(1)}, y^{(2)}, y^{(3)}, y^{(4)}, y^{(5)}\}
\]

To establish an upper-bound translation quality estimate, an \textbf{Oracle Rescoring} step was performed. Sentence-level BLEU scores (Papineni et al., 2002)\cite{papineni2002bleu} were computed between each candidate and the gold reference. The candidate achieving the highest BLEU score was selected as:
\[
y^{*} = \arg\max_{i} \, \text{BLEU}(y^{(i)}, y_{\text{ref}})
\]

This approach offers a robust evaluation by accounting for multiple plausible translations and selecting the best match.

%% file: chapters/chapter_5.tex
After training the BiLSTM baseline and fine-tuning the mT5 and NLLB-200 checkpoints, the performance of all model variants was evaluated on the reserved test set of 1,467 informal Bangla-English sentence pairs. During inference, the BiLSTM demonstrated the weakest performance, while the NLLB architectures showed superior generalization. The standard \textbf{Bilingual Evaluation Understudy (BLEU)} score, along with the assessment of training loss coupled with the more qualitative assessment of translations produced by the models, provided the basis for the evaluation.

\renewcommand{\thefigure}{5.\arabic{figure}}
\setcounter{figure}{0}

\subsection{Quantitative Performance Metrics}
The BLEU score served as the principal measure for inter-model comparison as it attempts to quantify the quality of translations by analyzing the overlap of $n$-grams of the machine translation and the corresponding human target (Papineni et al, 2002)\cite{papineni2002bleu}.

\subsubsection{BLEU Scores}
The performance of the models, analyzed and presented in Table~\ref{tab:bleu-score-final}, illustrates the progression of translation quality from basic RNNs to large-scale Transformers. A clear hierarchy of performance is evident, confirming the benefits of transfer learning and increased parameter capacity for this informal, low-resource task.

\begin{table}[htbp]
  \centering
  \caption{Comparative Quantitative Performance: BLEU Scores}
  \label{tab:bleu-score-final}
  \begin{tabularx}{\textwidth}{@{} l c c @{}}
    \toprule
    \textbf{Model Architecture} & \textbf{Validation Score} & \textbf{Test Score} \\
    \midrule
    BiLSTM + Attention Mechanism & 3.23 & 1.07 \\
    \midrule
    mT5 (small) & 14.91 & 11.69 \\
    mT5 (large) & 44.21 & 35.19 \\
    \midrule
    NLLB-200 (600M) & 42.68 & 42.37 \\
    NLLB-200 (1.3B) & 49.08 & 48.11 \\
    \textbf{NLLB-200 (3.3B)} & \textbf{57.42} & \textbf{56.83} \\
    \bottomrule
  \end{tabularx}
\end{table}

As anticipated for a limited-data task, the sequence-to-sequence BiLSTM model exhibited poor generalization, achieving a test BLEU score of only 1.07. On the other hand, the transformer-based models did perform comparatively better. Among the models we evaluated, the \textbf{NLLB-200} family consistently outperformed the mT5 variants. Even the distilled NLLB-600M model achieved a test score of 42.37, surpassing the significantly larger mT5-large (35.19). This suggests that NLLB's massively multilingual pre-training provides a stronger initialization for low-resource translation than the general text-to-text objective of mT5.

The best results among the fine-tuned models belonged to the \textbf{NLLB-200 (3.3B)} with QLoRA for fine-tuning, obtaining a test BLEU score of \textbf{56.83}, a large gain compared to the mT5-large. The BLEU score range of the different models within the NLLB family (42.37 $\rightarrow$ 48.11 $\rightarrow$ 56.83) is also further evidence for a positive scaling law, stating that an increase in model capacity enables the model to better grasp complex, irregular language phenomena in informal Bangla.

\subsubsection{Comparative Evaluation with State-of-the-Art LLMs}
To contextualize the performance of our fine-tuned NLLB-3.3B model, research conducted a comparative evaluation against \textbf{GPT-4o mini}, a leading proprietary Large Language Model (LLM). This comparison assesses whether a specialized, fine-tuned open-source model can compete with generalized industrial systems.

\begin{table}[htbp]
  \centering
  \caption{Comparison against State-of-the-Art Proprietary LLM}
  \label{tab:llm-comparison}
  \begin{tabularx}{\textwidth}{@{} l c c @{}}
    \toprule
    \textbf{Model} & \textbf{Method} & \textbf{Test Score (BLEU)} \\
    \midrule
    \textbf{NLLB-200 (3.3B)} & Fine-Tuned (QLoRA) & 56.83 \\
    GPT-4o mini & API implementation & \textbf{58.15} \\
    \bottomrule
  \end{tabularx}
\end{table}

As shown in Table~\ref{tab:llm-comparison}, the fine-tuned NLLB-3.3B model achieved performance nearly at parity with GPT-4o mini (56.83 vs. 58.15). While the proprietary model holds a slight edge, our result demonstrates that efficiently fine-tuning moderate-sized open-source models (3.3B) can yield professional-grade translation quality comparable to massive commercial systems (~8B+ parameters), without the associated inference costs or dependency on external APIs.

\subsubsection{Loss Analysis}
An analysis of the training and validation loss curves provided an insight into the learning dynamics of each architecture, in particular with regard to convergence and risk of immediate overfitting. \\[5pt]

\begin{figure*}[htbp]
  \centering

  \begin{minipage}[t]{0.32\textwidth}
    \centering
    \includegraphics[width=\linewidth]{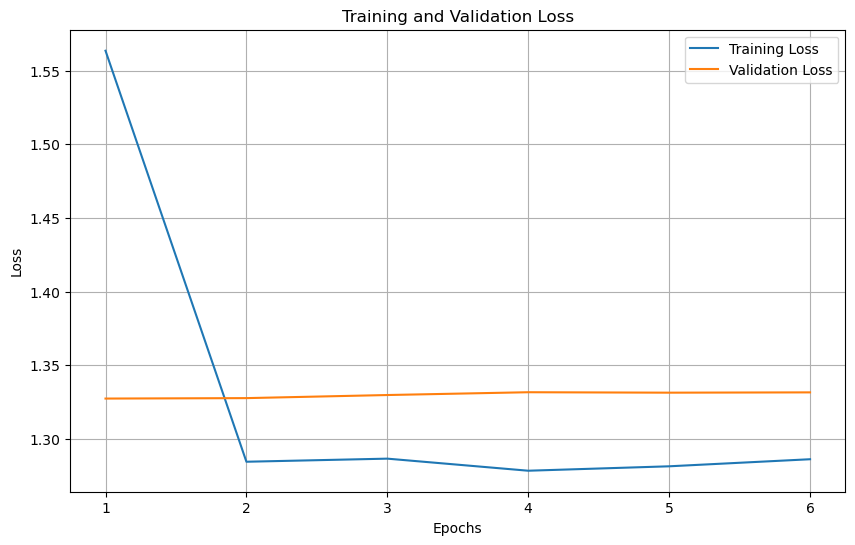}
    \caption{BiLSTM Loss Graph}
    \label{fig:bilstm}
  \end{minipage}\hfill
  \begin{minipage}[t]{0.32\textwidth}
    \centering
    \includegraphics[width=\linewidth]{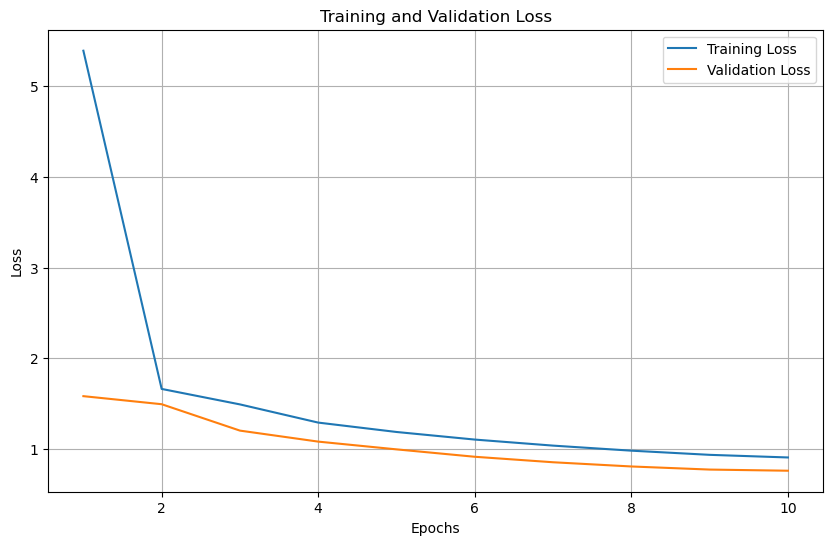}
    \caption{mT5-small Loss Graph}
    \label{fig:mt5-small}
  \end{minipage}\hfill
  \begin{minipage}[t]{0.32\textwidth}
    \centering
    \includegraphics[width=\linewidth]{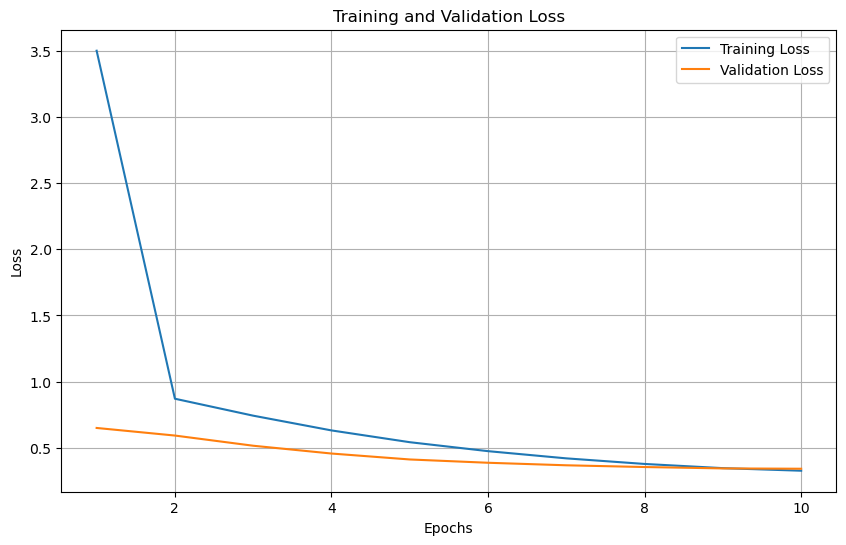}
    \caption{mT5-large Loss Graph}
    \label{fig:mt5-large}
  \end{minipage}

  \vspace{0.4cm}

  \begin{minipage}[t]{0.32\textwidth}
    \centering
    \includegraphics[width=\linewidth]{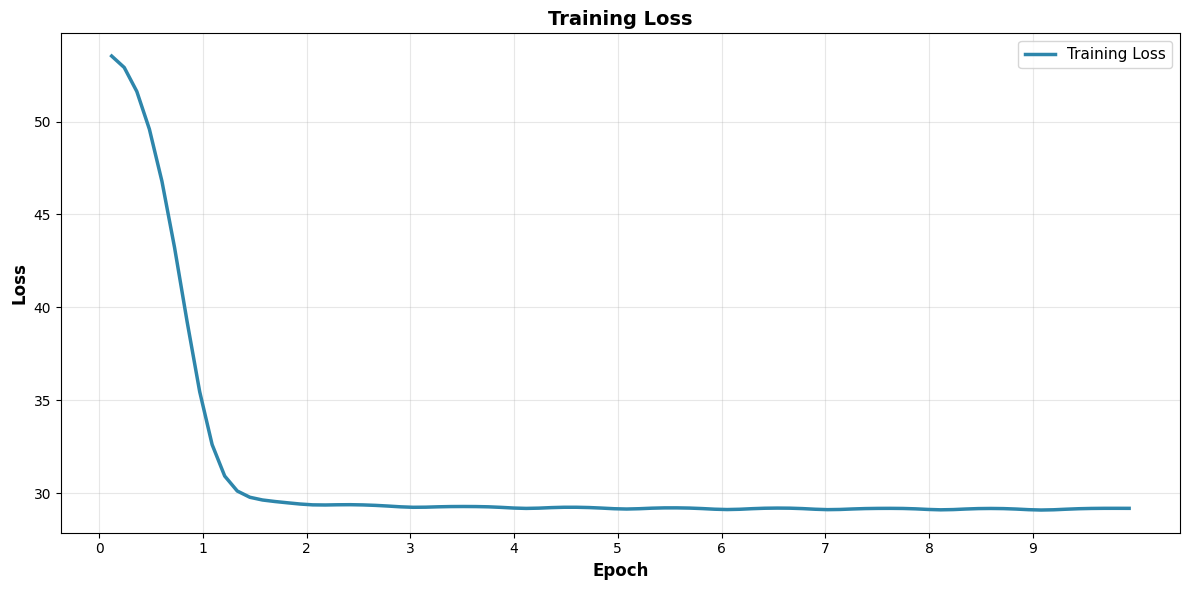}
    \caption{NLLB-200 (600M) Loss}
    \label{fig:nllb1}
  \end{minipage}\hfill
  \begin{minipage}[t]{0.32\textwidth}
    \centering
    \includegraphics[width=\linewidth]{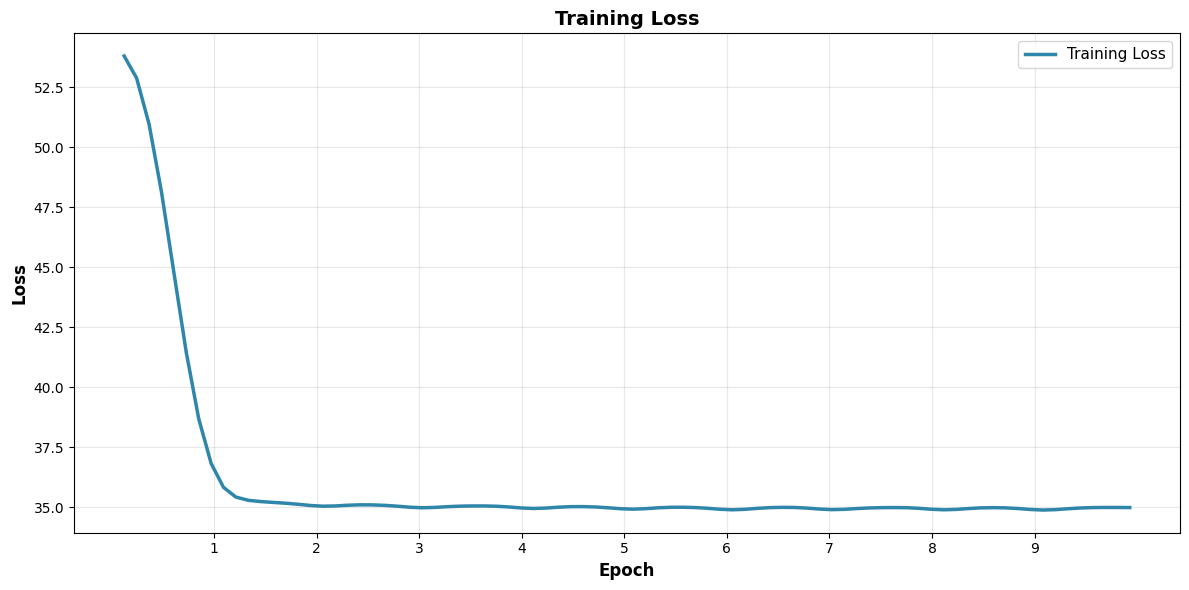}
    \caption{NLLB-200 (1.3B) Loss}
    \label{fig:nllb2}
  \end{minipage}\hfill
  \begin{minipage}[t]{0.32\textwidth}
    \centering
    \includegraphics[width=\linewidth]{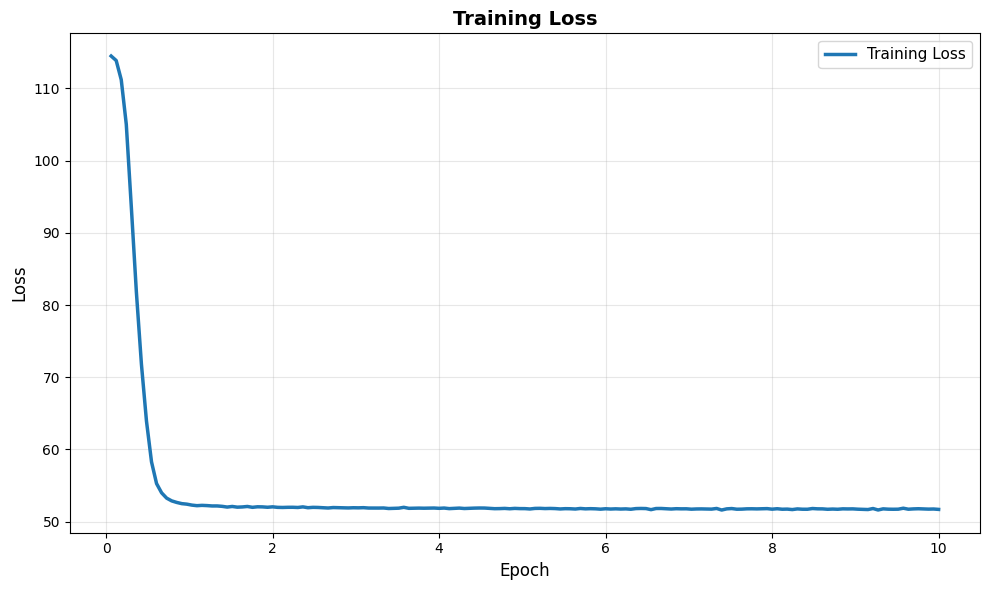}
    \caption{NLLB-200 (3.3B) Loss}
    \label{fig:nllb3}
  \end{minipage}

  \caption*{Loss curves of different models used in the experiments}
\end{figure*}

The BiLSTM model \ref{fig:bilstm} demonstrated signs of immediate over-fitting. The validation loss remained constant or increased slightly after only 5 epochs, which led to an early stopping to avoid further divergence from the test data. This performance confirms that the architecture, despite including a focus mechanism, lacks the capacity to generalize from a limited amount of interview data. \\[5pt]

On the other hand, the transformer variants (mT5 and NLLB) showed good convergence profiles. For the mT5 variants, the loss curves shown in \ref{fig:mt5-small} and \ref{fig:mt5-large} depicted a consistent and steady decrease over 10 epochs. Even if the training process demonstrated some increases in validation loss in relation to training loss, particularly towards the end, the overall trend of convergence was positive and stable. Similarly, the NLLB models exhibited rapid loss reduction in \ref{fig:nllb1}, \ref{fig:nllb2}, and \ref{fig:nllb3} during the early epochs followed by stable convergence, indicating efficient adaptation despite their substantially larger parameter sizes. Notably, the superior BLEU scores of the NLLB models suggest they achieved the most efficient adjustment to the informal domain.

\vspace{0.2cm}
\subsection{Qualitative Analysis of Model Predictions}
A comprehensive qualitative analysis was conducted to evaluate the translation quality of each model architecture, using a representative set of informal Bangla-English sentence pairs. Table~\ref{tab:qualitative-comparison} provides a side-by-side comparison of outputs from the BiLSTM, mT5-small, mT5-large, NLLB-600M, NLLB-1.3B, and NLLB-3.3B models. This analysis showcases the challenges and capabilities of each model for handling low-resource, informal language.

\begin{sidewaystable}[p]
\centering
\caption{Qualitative Comparison of Model Predictions}
\label{tab:qualitative-comparison}
\renewcommand{\arraystretch}{1.15}
\footnotesize

\begin{tabularx}{\linewidth}{@{} p{2.8cm} X X X X X X X @{}}
\toprule
\textbf{Input (Bangla)} &
\textbf{Target (English)} &
\textbf{BiLSTM} &
\textbf{mT5-small} &
\textbf{mT5-large} &
\textbf{NLLB-600M} &
\textbf{NLLB-1.3B} &
\textbf{NLLB-3.3B} \\
\midrule

{\bengalifont কেমন হইছে জানাবেন} &
Do let me know how it was &
has the been been the &
why do you know how do you know what is &
what’s going on let us know what happened &
let me know how it be &
let me know how it is &
let me know how it went \\

{\bengalifont নাটকটা কি এখন দেখা যাবে ইউটিউবে} &
Can I see the drama on YouTube now &
the the the the the &
where will you see the drama of dramas and dramas will you see how will you &
do you see the drama which drama will be shown on the youtube &
can the drama be seen on youtube now &
can the drama be seen on youtube now &
can the drama be seen now on youtube \\

{\bengalifont বেশি না, এমন একটা বাশ খাবেন} &
Not that much, you will suffer &
not of the of to &
i do not eat such a bottle but i will
eat such a pot &
it would not be too much to eat a curry that is why you will &
not too much eat such a mushroom &
not much you will eat such a bamboo &
not too much you will eat such a bush \\

\bottomrule
\end{tabularx}
\end{sidewaystable}

\subsubsection{BiLSTM + Attention}
The results of the BILSTM + Attention model were very disappointing. As shown in Table~\ref{tab:qualitative-comparison}, the translations often degenerate into fragmented, repetitive strings of functional words, prepositions, and sentences ("the the the the the", "has the been been the"). These results were completely non-significant and confirmed the model's serious lack of generalization capability, which is directly correlated to its near-zero BLEU score of 1.07.

\subsubsection{mT5-small}

The mT5-small model produced generally readable outputs, but it did not understand the context and meaning of the source sentence, which is verified in Table~\ref{tab:qualitative-comparison}. Many of the translations also incorporated meaningless generics and poorly ordered structures, as with the second example, which provided a disorganized redundant answer composed of fragments.

\subsubsection{mT5-large}
With more parameters, the mT5-large model gaining the ability to produce more coherent and contextually accurate translations. Although the model produced accurate and fluent translations more often than not. At times it would generate direct literal translations that would sound unnatural, particularly with phrases that had a specific idiomatic use. Overall, it was able to produce translations with more cultural context than previously.
\subsubsection{NLLB-200 (600M)}

The distilled NLLB-600M model, despite being the smallest in the NLLB family, demonstrated a significant leap in coherence compared to the baseline models. As seen in Table ~\ref{tab:qualitative-comparison}, the predictions are syntactically correct and readable English sentences. However, the model still struggled with cultural nuances; notably, in the third example, it provided a literal translation of the Bangla idiom ("eat a mushroom") rather than capturing the implied meaning of suffering or getting into trouble.

\subsubsection{NLLB-200 (1.3B)}

Scaling up to the 1.3 billion parameter model yielded noticeable improvements in fluency and sentence structure. Table~\ref{tab:qualitative-comparison} illustrates that the model began to use more natural English phrasing (e.g., "let me know how it is"). While the overall translation quality improved, the model cannot capture the literal interpretation of the Bengali word {\bengalifont বাশ}, suggesting that 1.3B parameters may still be insufficient to fully resolve deeply colloquial metaphors without more specific idiomatic training data.

\subsubsection{NLLB-200 (3.3B)}
The largest model, NLLB-3.3B, demonstrated the highest level of sophistication, effectively bridging the gap between literal translation and contextual understanding. Table~\ref{tab:qualitative-comparison} highlights its superior performance: it not only generated perfectly fluent English for standard queries but, closely interpreted the figurative idiom in the third example ("you will eat such a bush") instead of a literal translation. This confirms that the massive parameter capacity, combined with efficient QLoRA fine-tuning, enabled the model to capture the deep semantic meaning of informal, culturally specific language.

In summary, while mT5-large produced contextually appropriate outputs, the significantly higher quantitative scores of the NLLB models suggest that the move towards massive multilingual architectures provides the necessary capacity to bridge these remaining gaps in idiomatic and cultural translation.

%% file: chapters/chapter_6.tex
The challenge of advancing machine translation (MT) for Bangla, a language used by millions but critically categorized as low-resource, is particularly pronounced in the domain of colloquial speech. This study successfully addressed this deficit by prioritizing the fundamental task of creating a specialized, manually curated, and augmented parallel corpus derived from social media and conversational text. We established the initial framework by systematically reviewing the extant literature on low-resource MT and then undertook the complex, labor-intensive process of data collection and translation. \\[5pt]\

The experimental results definitively validated our central hypothesis regarding architectural choice. Training the custom recurrent model, the BiLSTM with Attention, proved fundamentally insufficient for generalizing on this nuanced, data-constrained task, yielding a minimal performance score (Test BLEU: 1.07). Conversely, leveraging the large-scale pre-training knowledge inherent in the transformer paradigm was highly effective. Fine-tuning the \textbf{mT5-large} variant allowed us to not only achieve a remarkable test BLEU score of \textbf{35.19} but, more importantly, produce translations that were contextually appropriate and easy to read. The most notable outcome of this study is the superior performance of the \textbf{NLLB-200 family}, with its base variants already surpassing \textbf{mT5-large}. Moreover, parameter-efficient fine-tuning of the \textbf{NLLB-200 3.3B} model using \textbf{LoRA/QLoRA} produced the strongest results overall, achieving a \textbf{Test BLEU score of 56.83}. This improvement, both qualitatively and quantitatively, shows that transfer learning is the most appropriate approach for improving low-resource MT for complex, domain-specific cases, such as informal Bangla. This research has, first and foremost, addressed the refinement of advanced models and the informal dataset as a foundational step towards reducing the technological disparity for the language. \\[5pt]\

However, the pursuit of rigor requires acknowledging the limitations that define our future road-map. Even after increasing the initial 7,664 pairs, the final corpus size is still small, necessitating validation-based early stopping to prevent over-fitting. The human-centric augmentation and translation, done by people on the team who are not trained linguists, could raise some subjective biases, inaccuracies, and errors. Further, focusing on written social media posed a risk of excluding people who are not digitally active, which, in turn, may have limited the dataset’s representation of the sociolinguistic and dialectal spectrum. Most importantly, the best-performing model, NLLB-200 (3.3B), did not fully handle non-literal figurative language, a clear indication of the challenge posed by idioms on automatic translation. \\[5pt]\

To address these challenges, future work will focus on specific, more achievable goals. First, a focused initiative to grow our informal corpus in both size and dialect diversity is essential. For further research, we will share this resource, and it will be open source to promote collaborative efforts in the field. At the same time, we will work on next-generation architectures, targeting the higher-capacity Large Language Models, especially the \textbf{GPT family}, which may have enhanced intrinsic abilities regarding fine-tuning and zero-shot performance. Lastly, there is a need for a dedicated approach to address the failure modes currently in effect with idioms where the system ceases to move with literal phrase-for-phrase translation and instead captures the context of informal speech. We expect this research to be a starting point for closing the gap on digital accessibility for millions of Bangla speakers and enhancing the language's technical presence in the global arena.